\documentclass[sigconf]{acmart}

\usepackage{algorithm}
\usepackage{algorithmic}

\usepackage{amsmath,amsfonts,bm,nccmath}

\def\eqref#1{equation~\ref{#1}}

\def\1{\bm{1}}

\def\vx{{\bm{x}}}

\def\mA{{\bm{A}}}

\def\mE{{\bm{E}}}
\def\mF{{\bm{F}}}

\DeclareMathAlphabet{\mathsfit}{\encodingdefault}{\sfdefault}{m}{sl}
\SetMathAlphabet{\mathsfit}{bold}{\encodingdefault}{\sfdefault}{bx}{n}

\def\sR{{\mathbb{R}}}

\def\emA{{A}}

\newcommand{\E}{\mathbb{E}}

\DeclareMathOperator*{\argmin}{arg\,min}

\newcommand{\stitle}[1]{\vspace{1ex}\noindent{\bf #1}}

\usepackage{multirow}
\usepackage{latexsym}
\usepackage{url}
\usepackage{amsmath,nccmath}
\usepackage{xcolor}
\usepackage{graphicx}
\usepackage{subcaption}
\usepackage{caption}
\usepackage{enumitem}
\usepackage{array}

\AtBeginDocument{%
  \providecommand\BibTeX{{%
    \normalfont B\kern-0.5em{\scshape i\kern-0.25em b}\kern-0.8em\TeX}}}

\copyrightyear{2024}
\acmYear{2024}
\setcopyright{acmlicensed}\acmConference[KDD '24]{Proceedings of the 30th ACM SIGKDD Conference on Knowledge Discovery and Data Mining}{August 25--29, 2024}{Barcelona, Spain}
\acmBooktitle{Proceedings of the 30th ACM SIGKDD Conference on Knowledge Discovery and Data Mining (KDD '24), August 25--29, 2024, Barcelona, Spain}
\acmDOI{10.1145/3637528.3671829}
\acmISBN{979-8-4007-0490-1/24/08}

\newcolumntype{P}[1]{>{\centering\arraybackslash}p{#1}}

\newcommand{\suhang}[1]{{\color{blue}[SW: #1]}}

\newcommand{{\method}}{SelMAG}

\begin{document}

\title{Multi-source Unsupervised Domain Adaptation on Graphs with Transferability Modeling}

\author{Tianxiang Zhao}
\affiliation{%
  \institution{The Pennsylvania State University}
  \state{State College}
  \country{USA}
}
\email{tkz5084@psu.edu}

 \author{Dongsheng Luo}
\affiliation{%
  \institution{ Florida International University}
  \city{Miami}
  \country{USA}}
\email{dluo@fiu.edu}

\author{Xiang Zhang}
\affiliation{%
  \institution{The Pennsylvania State University}
  \state{State College}
  \country{USA}
}
\email{xzz89@psu.edu}

\author{Suhang Wang}
\affiliation{%
  \institution{The Pennsylvania State University}
  \state{State College}
  \country{USA}
}
\email{szw494@psu.edu}

\begin{abstract}
In this paper, we tackle a new problem of \textit{multi-source unsupervised domain adaptation (MSUDA) for graphs}, where models trained on annotated source domains need to be transferred to the unsupervised target graph for node classification. Due to the discrepancy in distribution across domains, the key challenge is how to select good source instances and how to adapt the model. Diverse graph structures further complicate this problem, rendering previous MSUDA approaches less effective. In this work, we present the framework Selective Multi-source Adaptation for Graph ({\method}), with a graph-modeling-based domain selector, a sub-graph node selector, and a bi-level alignment objective for the adaptation. Concretely, to facilitate the identification of informative source data, the similarity across graphs is disentangled and measured with the transferability of a graph-modeling task set, and we use it as evidence for source domain selection. A node selector is further incorporated to capture the variation in transferability of nodes within the same source domain. To learn invariant features for adaptation, we align the target domain to selected source data both at the embedding space by minimizing the optimal transport distance and at the classification level by distilling the label function. Modules are explicitly learned to select informative source data and conduct the alignment in virtual training splits with a meta-learning strategy. Experimental results on five graph datasets show the effectiveness of the proposed method.
\end{abstract}


\maketitle

\section{Introduction}

\begin{figure}[t]
    \centering
  \includegraphics[width=0.42\textwidth]{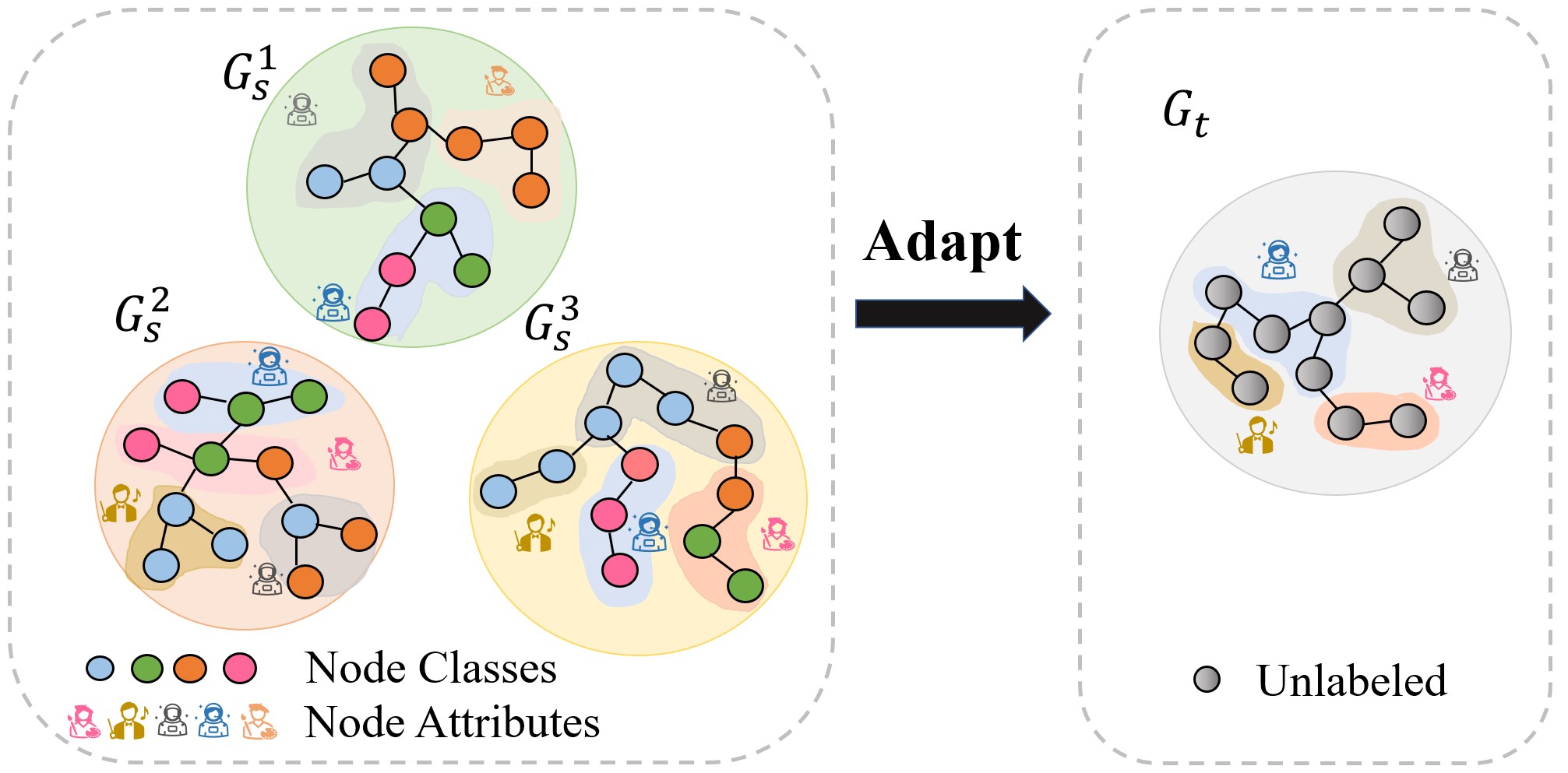}
  
    \vskip -1em
    \caption{An example of MSUDA for graphs. We want to transfer knowledge from annotated source domains $\{\bm{G}_s^1$, $\bm{G}_s^2$, $\bm{G}_s^3\}$ to $\bm{G}_t$. Regions of the same color denote similar node attributes. It can be observed that source domains and sub-graphs of each domain are of different importance in adapting to $\bm{G}_t$ w.r.t node distributions. }\label{fig:example}
    \vskip -2em
\end{figure}

Graph neural networks (GNNs) have shown great ability in representation learning on graphs, especially the node classification task~\cite{kipf2016semi,xu2018powerful}. Nevertheless, the success of GNNs heavily relies on label information; while for many real-world applications, obtaining label information is costly and time-consuming~\cite{ramalingam2018fake,jagabathula2022personalized}. The lack of label information challenges many existing GNNs. In practice, one often has access to multiple annotated domains in training. For example, social networks may be collected from different platforms, communities of different ethnicities, and users speaking different languages. For a newly-collected social network, typically node labels are unknown and we want to classify them with models trained from those source domains.  Recent years have featured a trend toward transferring knowledge across datasets to alleviate the lack of supervision~\cite{devlin2018bert,brown2020language}, which motivates us to explore the application of GNNs trained on annotated datasets (source domains) to the new unlabeled dataset (target domain)~\cite{Hamilton2017InductiveRL,DBLP:conf/www/WuP0CZ20} (the inductive setting). This scenario can be generalized to a new learning problem: \textit{multi-source unsupervised domain adaptation (MSUDA) for graphs} as shown in Fig.~\ref{fig:example}.




There are some significant challenges in solving the MSUDA task~\cite{sun2015survey,zhao2018adversarial}: (1) the mismatch between data distributions of the source and target domains, which requires an implicit or explicit adaptation of the trained model~\cite{blitzer2007learning,ben2006analysis}; (2) A clear discrepancy also exists among multiple source domains, hampering the effectiveness of mainstream single-source domain adaptation methods~\cite{DBLP:conf/nips/CourtyFHR17,DBLP:conf/cvpr/TzengHSD17}. Explorations have been made addressing these challenges. Existing
methods differentiate the effects of source domains on the target one by computing domain similarities from the lens of different perspectives, such as conditional distribution probability~\cite{DBLP:journals/tkdd/ChattopadhyaySFDPY12,sun2011two}, and estimation with an adversarial discriminator~\cite{zhao2018adversarial}, etc. Then, they can transform it into a single-source problem via re-weighting and conduct alignment for the adaptation~\cite{nguyen2021most}. Furthermore, ~\cite{ben2010theory,crammer2008learning,mansour2008domain} provide a performance bound of adapting from a weighted combination of source domains to the target domain.


However, all the aforementioned methods are designed for independent and identically distributed (i.i.d) data; while the complex graph structures bring new challenges to MSUDA. In graph-structured data, nodes are interconnected with edges, and the number of possible topology structures grows exponentially with the graph size. This rich and highly-diverse input space offers increased freedom for the mapping in <graph structure, node annotation> pairs. As a result, computing domain similarities in the input or embedding space and taking it as the estimation of informativeness for node classification could be unreliable, and it is unsafe to assume that nodes of the same source domain would carry a similar level of knowledge for this transfer, as the example in Fig.~\ref{fig:example}. It is difficult to identify subsets of source domains containing discriminative knowledge to transfer to the target domain due to the intricate nature of graphs. Furthermore, complex graph structures also rise difficulties in alleviating discrepancy among domains, rendering the conventional approach of aligning only in the embedding domain less-effective~\cite{DBLP:conf/nips/CourtyFHR17,long2015learning}. 


To address the aforementioned challenges,  we propose to comprehensively depict the similarity between graphs from multiple views, such as distribution of node attributes, edge existences, topological structures, etc. The informativeness of source graphs should further be modeled in aware of the downstream node classification task, and at both graph and node levels to prevent sub-optimal adaptations. As shown in \cite{blitzer2007learning}, the performance of domain adaptation is bounded by the discrepancy in distribution between the source and the target data set, and a tighter adaptation bound can be derived by identifying subsets of source data that are more similar to the target domain~\cite{crammer2008learning,ben2010theory}. With beneficial source data identified, we can achieve the adaptation through aligning the embedding learning and label prediction processes. 

Concretely, we design the first framework, {\method}, for MSUDA on graphs. First, a set of graph modeling tasks are selected to capture the graph distribution of each domain. The transferability of models trained on these tasks provides a numeric estimation of domain similarity from different perspectives. Then, based on obtained similarity measurements, a source-graph selector is designed to identify informative source domains towards the target domain on the target downstream task. Furthermore, a sub-graph node selector is adopted to assign different important scores to nodes of the same source domain. It can help conduct more fine-grained selections and meanwhile alleviate the target shift problem (like when class proportions are different)~\cite{redko2019optimal}. Virtual training splits are designed to explicitly optimize these selectors in capturing transferability across graph domains with Meta-learning~\cite{finn2017model}. Finally, an optimal-transport-based alignment is conducted both in the embedding space and the classification space. 
Our main contributions are: 
\begin{itemize}[leftmargin=0.2in]
\item We study a novel problem of MSUDA for graph-structured data, by selecting informative source data and adapting in both the embedding and classification space.
\item To identify sub-graphs of source domains that can transfer discriminative knowledge, we design a modeling-based graph selector and a sub-graph node selector, and explicitly train them on the selection-and-adapt pipeline with meta-learning.
\item We design an adaptation objective by conducting bi-level alignments with optimal transport and knowledge distillation, which can intrinsically incorporate the learned informativeness of source domains simultaneously.
\item Experimental evaluations show that {\method} achieves state-of-the-art performance on five datasets, validating its design. Case studies further show the ability of {\method} in capturing the informativeness of source domains. 
\end{itemize}

\section{Related Work}

\subsection{Graph Neural Network}
With the increasing need for learning on relational data structures~\cite{Fan2019GraphNN,dai2022towards}, various graph neural networks have been designed, 
which can be categorized into spectral approaches ~\cite{bruna2013spectral,Tang2019ChebNetEA,kipf2016semi} and spatial approaches~\cite{Hamilton2017InductiveRL,atwood2016diffusion,xiao2021learning}.  Despite their differences, most GNNs fit within the framework of message-passing~\cite{gilmer2017neural}, in which nodes are iteratively updated by aggregating messages from local neighborhoods.  For instance, GCN~\cite{kipf2016semi} passes messages from neighboring nodes with fixed weights, GAT~\cite{velivckovic2017graph} applies the self-attention mechanism to learn different attention scores and selects neighborhood messages dynamically. Some works\cite{lin2022prototypical,ahmadi2020memory,xu2022hp} augment GNNs with explicit prototypes to hierarchically model the motif structures and increase the data efficiency. Other works~\cite{yang2020factorizable,zhao2022exploring,xiao2022decoupled} propose to uncover latent groups of nodes or edges and pass messages on the disentangled graph. 
Recently, explorations have also been made over the trustworthiness of graph neural networks~\cite{dai2022comprehensive,wu2022survey,ren2022mitigating,ren2022semi} and their explainability~\cite{dai2021towards,zhao2022consistency}. 

Despite the great success of GNNs, their success usually hinges upon the availability of labeled training data, especially for the task of node classification~\cite{kipf2016semi,zhao2024disambiguated}. However, in practice, for a newly-forming or low-resource target graph, its nodes are usually unlabeled. The lack of labeled data challenges many existing GNN classifiers. Thus, we often need to adapt a model trained from mature source domains which have abundant label information to the target domain.  However, distribution shifts often exist between the source domains and the target domain, which calls for the development of domain adaptation algorithms on graph~\cite{DBLP:conf/www/WuP0CZ20}. Particularly, as there are usually multiple labeled source domains can be exploited, we propose to study a novel problem of multi-source domain unsupervised adaptation on graphs, which aims to adapt classifiers from multiple source graphs to the target unlabeled graph.

\subsection{Unsupervised Domain Adaptation}

Unsupervised domain adaptation (UDA) aims to transfer the knowledge learned from labeled source domains to the unlabeled target domain~\cite{blitzer2007learning,ben2006analysis,wilson2020survey,rentablog}. To adapt the knowledge from the source domains to the target domain, one major challenge is how to address the distribution shift between source and target domains. Most recent UDA methods focus on aligning source and target domains by learning domain-invariant features~\cite{DBLP:conf/cvpr/TzengHSD17,fernando2013unsupervised}, encouraging features to follow the same distribution regardless of which domain they come from~\cite{wilson2020survey}. To reduce the distribution discrepancy, some works~\cite{long2015learning,DBLP:conf/cvpr/YanDLWXZ17,ren2023t} propose to minimize a divergence
that measures the distance between distributions. \cite{ghifary2016deep,bousmalis2016domain} propose to learn representations that can reconstruct the data distribution of the target domain. Other works~\cite{DBLP:conf/cvpr/TzengHSD17,DBLP:conf/nips/CourtyFHR17} use adversarial learning to fool a discriminator which is trained to differentiate between two distributions.  
The aforementioned approaches are all designed for i.i.d data. Recently, there are few works on domain adaptation on other data structures, e.g., graph data~\cite{zhang2019dane,mao2021source}, text data~\cite{chen2009extracting,zhang2019sequence,xiao2022domain,ren2024analyzing,ren2024esacl}, etc. For example, UDA-GCN~\cite{DBLP:conf/www/WuP0CZ20} extends adversary alignment~\cite{DBLP:conf/cvpr/TzengHSD17} to graphs with an attentive feature extractor to learn the invariant features.
Overall, it is non-trivial to formulate graph distribution and design adaptive models considering its highly-diverse topology, graph size, and node features \cite{DBLP:conf/www/WuP0CZ20,wu2022non}. 

Particularly, this work is related to multi-source unsupervised domain adaptation. 
This task is complicated further by distribution discrepancy among source domains\cite{li2021dynamic,yang2020curriculum,peng2019moment,chen2023unsupervised}. Theoretical analyses have been provided w.r.t the performance bound of multi-source unsupervised domain adaptation~\cite{ben2010theory,crammer2008learning,mansour2008domain}, showing the importance of selecting important source domains. Explorations have been made in measuring domain similarities with conditional distribution probability from the smoothness assumption~\cite{DBLP:journals/tkdd/ChattopadhyaySFDPY12,sun2011two}. ~\citet{zhao2018adversarial} uses an adversarial discriminator and conducts the worst-case alignment, and ~\citet{nguyen2021most} adopts a model-based similarity estimation. However, all these methods are designed for i.i.d data and have difficulty in applying to graphs. In this work, addressing the highly-diverse graph structures, we propose to design a domain selector based on disentangled similarity measurements with graph-modeling tasks and further conduct sub-domain selection with a node-level selector, which is better in estimating informativeness of source domains w.r.t node classification of the target graph.

\section{Preliminary}
\subsection{Notations and Problem Definition}
\textbf{Semi-supervised Node Classification.}
We focus on the node-level classification task in this work. Specifically, we use $\mathbf{G} = (\mathcal{V},\mathcal{E}; \mF, \mA )$ to denote a graph, where $\mathcal{V}$ is the node set and $\mathcal{E} \subset \mathcal{V} \times \mathcal{V}$ is the set of edges. Nodes are accompanied by an attribute matrix $\mF \in \sR^{|\mathcal{V}| \times d}$, and $i$-th row of $\mF$ is the $d$-dimensional attributes of the corresponding node. $\mathcal{E}$ is described by an adjacency matrix $\mA \in \sR^{|\mathcal{V}| \times |\mathcal{V}|}$. $\emA_{vu}=1$ if there is an edge between node $v$ and $u$; otherwise, $\emA_{vu}=0$. $\bm{Y} \in \sR^{|\mathcal{V}|}$ is the class information for nodes in $\mathbf{G}$, obtained with an unknown labeling function $f$, and $R(\bm{Y})$is the number of classes. During training, only a subset of $\bm{Y}$, $\bar{\bm{Y}}$ is available, containing the labels for the training node set. Based on those labeled nodes, a hypothesis model $h$ is trained to recover the unknown function $f$ and to predict node classes: 
\begin{equation}
    h(v; \mathbf{G}) \rightarrow \bm{Y}_v, \quad \forall v \in \mathcal{V}.
\end{equation}

\noindent\textbf{Multi-source Unsupervised Graph Adaptation.}
We use \textit{domain} to define a distribution over the graph generation and its latent labeling function, denoted as $\langle D, f \rangle$. In this multi-source graph adaptation, we have partially labeled graphs collected from $K$ source domains as $\mathcal{G}_s = \{\mathbf{G}_s^k, \bar{\bm{Y}}_{s}^k\}_{k=1}^K$ and an unsupervised graph $\mathbf{G}_t$ from the target domain. Each source graph $\mathbf{G}_s^k$ is generated following the distribution $D_s^k$ and its $\bar{\bm{Y}}_{s}^k$ is obtained with latent labeling function $f_s^k$. The objective is to build a hypothesis classifier $h_t$ that works well on the target domain $\langle D_t, f_t \rangle$, predicting classes of nodes in $\mathbf{G}_t$ accurately. Concretely, the task can be formalized as:

\vspace{0.5em}
\noindent{}\textit{Given $K$ partially labeled graphs {$\{\mathbf{G}_s^k, \bar{\bm{Y}}_{s}^k\}_{k=1}^K$} from different source domains and an unsupervised graph $\mathbf{G}_t$ from the target domain $\langle D_t, f_t \rangle$, we aim to train a node classification model $h_t$ to simulate $f_t$ with a small loss $\mathcal{L}(f_t(\mathbf{G}_t), h_t(\mathbf{G}_t))$ on the target graph $\mathbf{G}_t$.
}

\subsection{Optimal Transport for DA}\label{sec:OT}
Optimal Transport (OT) provides a theoretic tool for computing distances between probability distributions and for alignment-based domain adaptation~\cite{DBLP:conf/nips/CourtyFHR17}, on which our method {\method} is developed. In this section, we provide an introduction to its key concepts and some important results used in the following sections. The OT problem searches for a plan with the minimum cost to transform a distribution ${D}_s$ (over space $\Omega_s$) to another distribution ${D}_t$ (over space $\Omega_t$). The cost of transforming each element is measured by a cost function $c: \Omega_s \times \Omega_t \rightarrow \mathbb{R}^+$. Following the Monge mass transfer problem~\cite{bogachev2012monge}, the search of transport plan $T$ with minimum total transportation cost $C(T)$ can be expressed as: 
\begin{equation}\label{eq:OT_1}
\begin{aligned}
    \argmin_T C(T) &=  \int_{\Omega_s} c(\vx, T(\vx)) d{D}_s(\vx), \\
     \text{s.t.}  & \quad T\#D_s = D_t,
\end{aligned}
\end{equation}
where $T()$ denotes the transport plan and $D_s()$ means the distribution function in the source domain. $T\#D_s$ is the \textit{image measure} of $D_s$ by $T$~\cite{DBLP:conf/nips/CourtyFHR17}, another probability measure defined over $\Omega_t$ satisfying:
\begin{equation}
    T\#D_s(B) = D_s(T^{-1}(B)), \quad \forall \text{ Borel subset } B \subset \Omega_t
\end{equation}
This guarantees that $T$ is a transport map from $D_s$ to $D_t$. In practice, cost function $c$ is usually set as the squared Euclidean distance~\cite{DBLP:conf/nips/CourtyFHR17}. As shown in ~\cite{kantorovich2006translocation},
 a convex relaxation of Eq.~(\ref{eq:OT_1}) can be derived as:
\begin{equation}\label{eq:OT_2}
    \argmin_{\gamma \in \Pi} \int_{\Omega_s \times \Omega_t} c(\vx_s, \vx_t)d\gamma(\vx_s,\vx_t),
\end{equation}
where $\Pi$ is defined to be the set of all probabilistic couplings in $ P(\Omega_s \times \Omega_t)$ with marginals $D_s$ and $D_t$. $\gamma$ can be understood as a joint probability measure for the transportation plan.

To expose the connection between optimal transport and domain adaptation, following previous analysis~\cite{ben2006analysis,redko2019optimal}, the error bound of applying source models to the target domain can be summarized as:
\begin{equation}\label{eq:bound}
    \epsilon_t ({h}) \leq \sum_{k=1}^K \alpha_k \cdot \epsilon_k(h) + \lambda^* + d_{\mathcal{H}}(\sum_{k=1}^K \alpha_kD_k ,D_t),
\end{equation}
where the first term represents errors on source domains, $\lambda^*= \min_h\{\epsilon_t(h)+\sum_{k=1}^{K}\alpha_k\epsilon_k(h) \}$ is the optimal joint error and is a constant, while $d_{\mathcal{H}}(\sum_{k=1}^K \alpha_kD_k, D_t)$ measures $\mathcal{H}$-divergence across two distributions. To achieve a low error on the target domain, the discrepancy between the distribution of the target domain and that of source domains needs to be minimized (the third term), which can be achieved by aligning them in the embedding space, corresponds to 
learning a generalizable feature extractor across domains and minimizing the optimal transportation cost $\min_T C(T)$ on it~\cite{DBLP:conf/nips/CourtyFHR17}.


\section{Methodology}

\begin{figure}[t]
    \centering
  \includegraphics[width=0.46\textwidth]{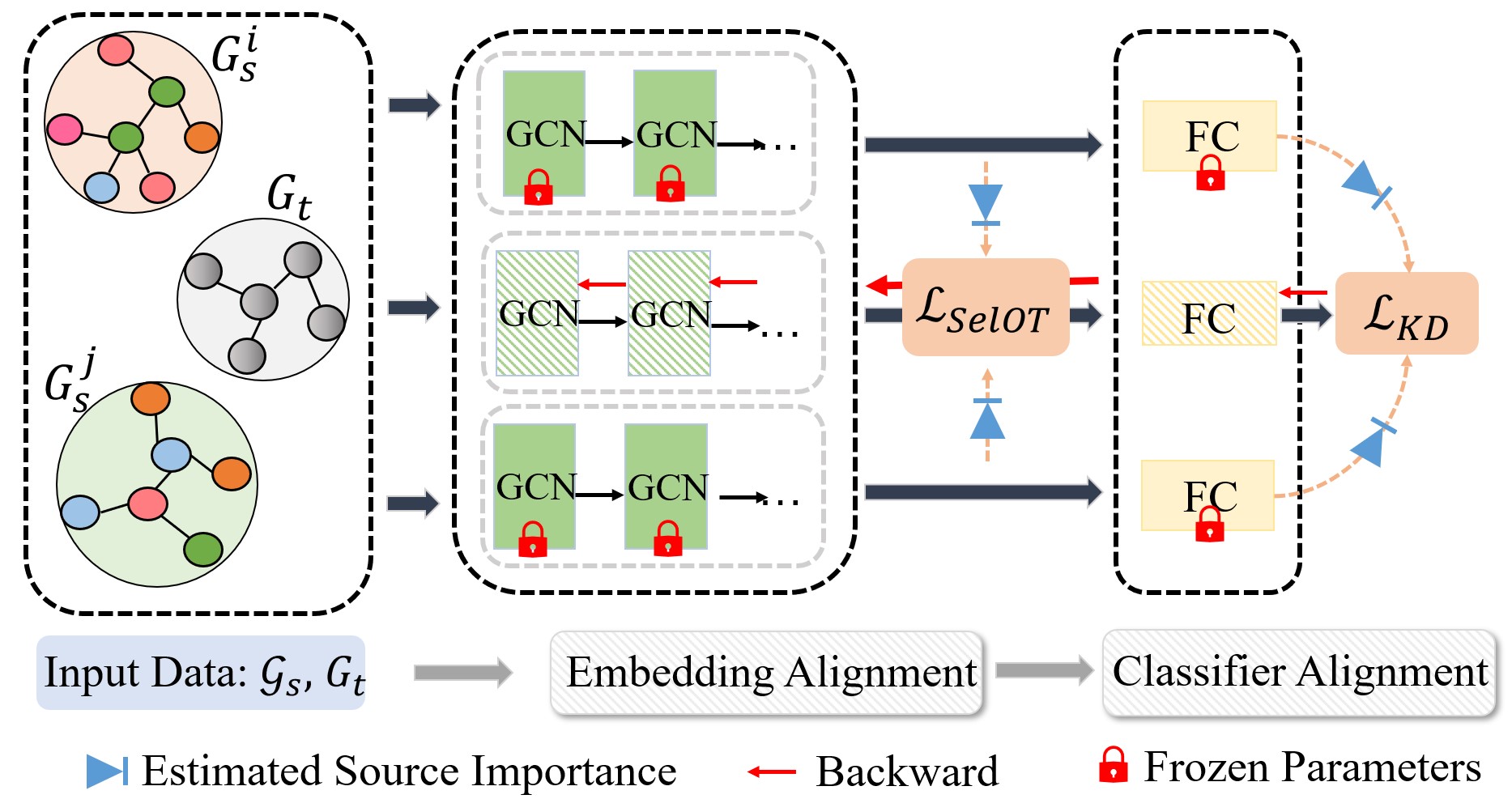}
  
    \vskip -1em
    \caption{Adaptation process of {\method}. Source importance is estimated with a global graph selector and sub-graph node selector, and is incorporated into this bi-level alignment.}\label{fig:framework}
    \vskip -1em
\end{figure}

In this section, we provide details of {\method}, which can identify informative subsets of source domains utilizing disentangled domain-level similarities together with sub-domain node-level selections. An overview of the adaptation process is provided in Fig.~\ref{fig:framework}.

\subsection{Learning of Source Models}

As we have labels for source domains, we can train a set of source models $\{h_s^k\}_{k=1}^K$, one for each domain, using the labeled data. Those models will be used to  train a target model $h_t$ later. Specifically, each hypothesis model $h$ is composed of two parts, a feature extraction module $g_{\text{ext}}$ (with stacked GNN layers) and a classifier module $g_{\text{cls}}$ (as an MLP). Taking node $v$ from $\mathbf{G}_s^i$ as an example, the feature extraction module first learns node representation of $v$ as:
\begin{equation}\label{eq:infer}
\begin{aligned}
    \vx_v = g_{\text{ext}}^i(v; \mathbf{G}_s^i),
\end{aligned}
\end{equation}
With the learned node representation $x_v$, the classifier module predict the label of node $v$:
\begin{equation}
    \hat{\bm{P}}_v = g_{\text{cls}}^i(\vx_v),
\end{equation}
$\hat{\bm{P}}_v \in \sR^{R(\bm{Y})}$ with each dimension representing the predicted probability of $v$ belonging to the corresponding class. For each source graph $\mathbf{G}_s^i$, cross-entropy loss is used for learning:
\begin{equation}
 \min_{h_s^i} \mathcal{L}_{CE}  = -\sum_{v \in \bar{\mathcal{V}}_i} \sum_{c=1}^{R(\bm{Y})}  \mathbf{1}(\bar{Y}_{v}==c) \cdot \log(\hat{P}_{v}[c]),
 \end{equation}
where $\bar{\mathcal{V}}_i$  denotes the labeled node set of $\mathbf{G}_s^i$. To promote the alignment in the embedding space, we adopt a hard parameter sharing on the feature extractor of all source domains and leave the classifier modules domain-specific.

The target model $h_t$ is learned upon two objectives: (1) aligning embeddings of the target graph to those of selected informative subsets of source graphs with a loss derived from optimal transport theory, and (2) the classification loss on pseudo labels generated with source models through a weighted distillation strategy. Details of them will be introduced in the following sections.

\subsection{Estimate Transferability  of Source Domains}
Distribution shifts result in different levels of transferability from source domains to $\mathbf{G}_t$, and sub-graphs of the same domain would also vary in importance for transferring discriminative features. 
To identify informative subsets of source graphs that can be transferred to the target domain, we adopt a coarse-to-fine paradigm. First, a graph-level selector is designed conditioned on factorized similarity measurements and then a node-level selector is proposed to capture informative subsets of each source graph. We will go into detail about them in this section.

\subsubsection{Modeling-based Graph Selector}\label{sec:graph_selector}
In this part, we introduce the design of the graph-level selector based on similarities that are factorized into the transferability of different graph modeling tasks.
Generally, if prediction models trained on graph $\mathbf{G}_i$ can perform well on $\mathbf{G}_t$, then $\mathbf{G}_i$ and $\mathbf{G}_t$ are similar in some ways. A set of self-supervision tasks can be designed to model the graph distribution from different perspectives, and we can train on $\mathbf{G}_s^k$ and test the performance on $\mathbf{G}_t$ to obtain disentangled similarity measurements. Various self-supervision tasks can be adopted. In this paper, we adopt the following three self-supervision tasks related to graph structure/properties, which can better model graph distribution for measuring the transferability. We leave the exploration of other tasks as future work.
\begin{itemize}[leftmargin=0.1in]
    \item Node Prediction~\cite{hu2019strategies}. In this task, we randomly mask a ratio of nodes of the graph as input, and the objective is to predict the attributes of those masked nodes.
    \item Edge Prediction~\cite{jin2020self}. We randomly sample a set of edges for training, and the objective is to predict the existence of edges for a random pair of nodes.
    \item Context Prediction~\cite{hu2019strategies}. Based on node attributes, we cluster the nodes of those graphs into multiple groups with the K-Means algorithm. Then, for each node, we obtain its so-called ``context'' as the group distribution of its direct neighborhood. The objective is to predict obtained context for each node.
\end{itemize}
It can be seen that these self-supervised learning (SSL) tasks focus on different aspects of the graph and have different reliance upon typical graph elements: nodes, edges, and neighborhood topology. \textit{The transferability of these tasks between each source graph and the target graph can depict their similarities from different perspectives}, which could be used to evaluate the informativeness of each source graph towards the downstream node classification task. 

We use $g_T^i$ to denote the prediction module w.r.t SSL task $T$ on graph $\mathbf{G}_i$, as shown in Fig.~\ref{fig:framework}. Concretely, first, we train modules $\{g_T\}_{T\in \mathcal{T}}$ w.r.t these graph modeling tasks (denoted as $\mathcal{T}$) for each graph respectively. Then, we quantify the transferability of these SSL tasks from graph $j$ to graph $t$ as the performance after exchanging the trained modules:
\begin{equation}
s_{T}^{j\rightarrow t} = \mathbb{E}_{v \in \mathcal{V}_t} \mathcal{L}_T(g_T^j(\vx_v),\mathbf{G}_{t}),
\end{equation}
where $\mathcal{V}_t$ is the node set of $\mathbf{G}_t$, $\vx_v$ is extracted node embedding following Eq.~\ref{eq:infer}, and $\mathcal{L}_T(g_T^j(\vx_v),\mathbf{G}_{t})$ measures loss of task $T$ for node $v$ of $\mathbf{G}_{t}$. $s_T^{j \rightarrow t}$ denotes the obtained performance of module $g_{T}^{j}$ when applied to $\mathbf{G}_t$. Finally, the graph-level selector $g_{\text{sel}}^{\text{global}}$ takes the transferred performance scores on those SSL tasks as input evidence and predicts the informativeness of graph $\mathbf{G}_j$ to $\mathbf{G}_t$ w.r.t the downstream task as:
\begin{equation}
\begin{aligned}
\hat{s}_{\text{global}}^{j \rightarrow t} &= g_{\text{sel}}^{\text{global}} \big(\{s_T^{j\rightarrow t}\}_{T\in \mathcal{T}}, \{s_T^{i\rightarrow t} \}_{T \in \mathcal{T}} \big)  \\
{s}_{\text{global}}^{j \rightarrow t} &= \frac{\hat{s}_{\text{global}}^{j \rightarrow t}}{\sum_{j \neq i} \hat{s}_{\text{global}}^{j \rightarrow t}},
\end{aligned}
\end{equation}
where the obtained $s_{\text{global}}^{j \rightarrow t} \in [0,1]$. Note we incorporate the performance $s_T^{i \rightarrow t}$ into the input, which can expose relative performance drop and help evaluate the distribution shift.

\subsubsection{Sub-graph Node Selector}
The modeling-based selector assigns a weight to each source graph encoding its importance as a whole. However, for each source graph, some of its sub-graphs may be more important than others in adapting to the target domain $\langle D_t, f_t \rangle$. To conduct a more fine-grained selection and identify informative node groups of each source graph, we further design a subgraph-level node selector in this part. Without loss of generality, we implement it as a $2$-layer MLP. For each candidate node $v$ of the source graph $\mathbf{G}_j$, we concatenate its embedding with the global representation of target graph $\mathbf{G}_t$ as the input:
\begin{equation}\label{eq:local_weight}
\begin{aligned}
  s_{\text{local}}^{j \rightarrow t, v} = g_{\text{sel}}^{\text{local}} \big( \vx_v^j, \text{pooling}_{u \in \mathbf{G}_t}(\vx_u^i) \big),
\end{aligned}
\end{equation}
where $\vx_v^j$ denotes the embedding of node $v$ from graph $\mathbf{G}_j$ obtained with its feature extractor following Eq.~\ref{eq:infer}. We take pooled node embeddings of $\mathbf{G}_j$ as its graph-level representation. Specifically, we adopt both max-pooling and mean-pooling and concatenate them together to preserve both distinct parts and global patterns of $\mathbf{G}_t$~\cite{zhao2023faithful}. This selector will give similar weights to source data with similar embeddings in effect.

However, it is challenging to learn these two selectors with back-propagation on the adaptation performance due to the unsupervised target domain. Hence, we propose to optimize them with meta-level updates, which will be introduced in Sec.~\ref{sec:meta}.

\subsection{Alignment-based Domain Adaptation}
With transferability between source graphs and the target graph encoded, in this section, we introduce our strategy to incorporate it into the adaptation process to train the classifier that works for $\mathbf{G}_t$.  Previous analysis shows that the error of cross-domain adaptation is bounded by both the global divergence across two domains and the class-wise distribution shifts (which can be measured as the optimal joint error)~\cite{ben2006analysis,sun2015survey}. Therefore, we design the learning objectives of model $h_t$ by mapping nodes of $\mathbf{G}_t$ into the same space as selected data of source domains for aligning embeddings and imitating classification behavior of selected source models for aligning the labeling functions. Specifically, we design an optimal-transport-based algorithm that is able to utilize extracted transferability intrinsically along with a weighted knowledge distillation mechanism. Details are provided in the following parts.

\subsubsection{Selective Optimal Transport for Adaptation}
Based on the analysis in Sec.~\ref{sec:OT} and the error bound presented in Eq.~\ref{eq:bound}, adapting the GNN model requires minimizing the distance between the target and source distributions, which can be achieved by reducing the minimum total transportation cost in Eq.~\ref{eq:OT_2}. To obtain a smoother transport plan and increase the optimization efficiency~\cite{cuturi2013sinkhorn}, an entropy-based regularization $NE$ is added and the alignment loss can be formulated as:
\begin{equation}\label{eq:OT_reg}
    \min_{\gamma} \mathcal{L}_{\text{OT}} = \int_{\Omega_s \times \Omega_t} c(\vx_s, \vx_t)d\gamma(\vx_s,\vx_t) + \epsilon NE(\gamma),
\end{equation}
where $NE(\gamma) = \int_{\Omega_s \times \Omega_t}\gamma(\vx_s, \vx_t)\log\gamma(\vx_s, \vx_t) d\vx_{s}d\vx_{t}$ is the negentropy of transport plan $\gamma$. This loss can guide the adaptation of the target model to reduce the optimal transport distance between the embedding of source graphs and that of the target graph.

However, this objective neglects the difference in informativeness among source instances toward the target data. In multi-source graph adaptation, different source graphs and sub-graphs of the same graph may contribute differently to the learning of the target model. Addressing this problem, we incorporate the predicted transferability $s_{\text{global}}$ and $s_{\text{local}}$ into the selective OT-based adaptation objective by augmenting the transport cost. Our basic idea is that the alignment should be focused on source data that have high importance for knowledge transfer to the target graph, and contrarily for the rest of the source data to prevent the problem of negative transfer. Hence, for node $v$ from source graph $\mathbf{G}_j$, the transport cost measurement w.r.t target graph $\mathbf{G}_t$ can be calculated as:
\begin{equation}\label{eq:weight}
    c^{\text{sel}}(\vx_v, \vx_t) = c(\vx_v, \vx_t) \cdot s^{j \rightarrow t }_{\text{global}} \cdot s^{j \rightarrow t, v}_{\text{local}},
\end{equation}
$c^{\text{sel}}(\vx_v, \vx_t)$ is used to replace the original cost measurement $c(\vx_s, \vx_t)$ in Eq.~\ref{eq:OT_reg}. This design enables us to select and highlight informative parts of the source graph set during the embedding alignment for model adaptation. For tractable optimization, we can obtain its dual form with the Fenchel-Rockafellar theorem~\cite{cuturi2013sinkhorn}:
\begin{equation}\label{eq:ot_cost}
\begin{aligned}
    \mathcal{L}_{\text{SelOT}} &= \max_{\beta} {\big\{} \int_{\Omega_s}\beta(\vx_s)d\vx_{s} + \int_{\Omega_t}\beta^{c}_{\epsilon}(\vx_t)d\vx_t {\big\}} \text{ where } \\ 
    \beta^{c}_{\epsilon}(\vx_t) & :=  \begin{cases}
    \min_{\vx_s} \{c^{\text{sel}}(\vx_s, \vx_t)-\beta(\vx_s)\}, \quad \epsilon=0 \\
    -\epsilon \log {\big(}\E_{\vx_s}[exp{\frac{\beta(\vx_s)-c^{\text{sel}}(\vx_s,\vx_t)}{\epsilon}}] {\big)}, \quad \epsilon >0
    \end{cases},
\end{aligned}
\end{equation}
in which $\beta()$ is a scoring function and can be simulated with a network. We set $\epsilon$ as a small positive number following~\cite{nguyen2021most}. 

\subsubsection{Weighted Knowledge Distillation}
As the domain adaptation performance is bounded both by the shifts on the embedding space and the shifts on the learned labeling functions~\cite{nguyen2021most}, to transfer the classification information from source models, we further adopt a knowledge distillation loss to provide training signals in the label space. Concretely, based on the importance score $s_{\text{global}}$ of each source model $h_{s}^k$, the soft pseudo label of node $v$ in $\mathbf{G}_t$ can be obtained as
\begin{equation}
    \bar{\bm{Y}}_{t,v} = \sum_{\mathbf{G}_j \in \mathcal{G}_s} h_s^k(v; \mathbf{G}_t) \cdot s_{\text{global}}^{j \rightarrow t}
\end{equation}
Then the target model $h_t$ can be trained as
\begin{equation}\label{eq:kd}
\begin{aligned}
    \mathcal{L}_{\text{KD}} = \E_{v \in \mathbf{G}_t}-\sum_{i=1}^{R(\bm{Y})} \bar{\bm{Y}}_{t,v}[i]\log h_t(v; \mathbf{G}_t)[i],
\end{aligned}
\end{equation}
where $R(\bm{Y})$  is the number of classes, same as the dimension of $\bar{\bm{Y}}_{t,v}$. $h_t(v; \mathbf{G}_t)$ is the output of target model $h_t$ for node $v$ of $\mathbf{G}_t$, with its $i$-th dimension as predicted probabiliy of falling into class $i$. This cross-entropy loss will distill the knowledge of learned labeling functions of source domains to the target domain. 

\subsubsection{Objective Function of {\method}}
Putting everything together, the final learning loss for the alignment-based domain adaptation can be written as:
\begin{equation}\label{eq:full_obj}
    \min_{h_t} \lambda \cdot \mathcal{L}_{\text{SelOT}} + (1-\lambda) \cdot \mathcal{L}_{\text{KD}},
\end{equation}
where $\lambda$ controls the balance between alignment on the embedding space and that on the labeling function space respectively.

\subsection{Optimization with Meta-learning}\label{sec:meta}
The adaptation objective in Eq.~\ref{eq:full_obj} emphasizes reducing the alignment distance. Directly optimizing the proposed selectors on this domain adaptation task is improper and may result in trivia solutions, as it provides no signals over transferring class-discriminative information and has the danger of highlighting uniformly distributed but non-informative (sub-)graphs. To improve the selective knowledge transfer for domain adaptation, we design a meta-learning-based optimization strategy~\cite{finn2017model} by simulating the unsupervised domain adaptation setting and guiding the learning of selectors based on performance after adaptation. This ``learning to learn'' pipeline can provide explicit learning signals for selectors.

Concretely, to make the meta-training process in conformity with the adaptation process and directly optimize it, we propose to directly \textit{learn to select from source graphs} by iteratively conducting two learning steps, inner update, and outer update. Below, we will show how these two steps are designed.

\stitle{Inner Update.} To guarantee consistency, setting on the inner update is designed to be also a multi-source graph adaptation task. For available source graph set $\mathcal{G}_{s}$, in each iteration, we sample a pseudo target graph $\hat{\mathbf{G}}_t$ and use its complement as pseudo source graphs $\hat{\mathcal{G}}_{s}$. The target model $h_t$ is updated for $\mathbf{T}$ steps on $\hat{\mathbf{G}}_t$ following Eq.~\ref{eq:full_obj}, simulating the process of adaptation. At $k$-th step, parameters are updated with:
\begin{equation}~\label{eq:inner_update}
    \theta_{t}^{k} = \theta_{t}^{k-1} - \alpha \cdot \nabla_{\theta_{t}^{k-1}} (\lambda \cdot \hat{\mathcal{L}}_{\text{SelOT}} + (1-\lambda)\cdot \hat{\mathcal{L}}_{\text{KD}}),
\end{equation}
where $\theta_t$ is the parameter of model $h_t$ and $\alpha$ is the optimization step size. Losses $\hat{\mathcal{L}}_{\text{SelOT}}$ and $\hat{\mathcal{L}}_{\text{KD}}$ are calculated on the pseudo source and target graphs.

\stitle{Outer Update.} Nodes of $\hat{\mathbf{G}}_t$ are labeled and can be utilized to evaluate the performance of graph adaptation. Gradients from adapted $h_t$ can be back-propagated to the weights $\hat{s}_{\text{global}}$, $\hat{s}_{\text{local}}$ of pseudo source graphs $\hat{\mathcal{G}}_s$, and further, be utilized to optimize the two selectors. Concretely, $\theta_{\text{sel}}$ can be optimized accordingly to explicitly improve upon ``learning to select'' as:
\begin{equation}~\label{eq:outer_update}
    \theta_{\text{sel}} = \theta_{\text{sel}} - \eta \cdot \sum_{v \in \hat{\mathbf{G}}_{i}} \nabla_{\theta_{\text{sel}}}\mathcal{L}_{\text{cls}}(v; \hat{\mathbf{G}}_t),
\end{equation}
where $\theta_{\text{sel}}$ is the parameter of model $g_{\text{sel}}^{\text{global}}$ and $g_{\text{sel}}^{\text{local}}$, $\eta$ is the learning rate, and $\mathcal{L}_{\text{cls}}$ is node classification loss on labeled nodes of $\hat{\mathbf{G}}_t$, with model parameters learned from inner update. 

These two steps are performed iteratively and can be summarized in Alg.~\ref{alg:Framework}. With inner and outer updates, both selectors are trained to predict informative subsets of source graphs for the target graph w.r.t the downstream node classification task. Therefore, they can be trusted to be applied to $\mathbf{G}_t$ afterward this meta-training phase. Following meta-training steps, in adapting to $\mathbf{G}_t$, we will fine-tune the parameter of $h_t$ following Eq.~\ref{eq:full_obj}, as shown in Alg.~\ref{alg:Framework}.


\begin{algorithm}[t!]
  \caption{Full Training Algorithm}
  \label{alg:Framework}
  \begin{algorithmic}[1] 
  \REQUIRE 
    $\mathbf{G}_{t} = (\mathcal{V}_t,\mE_t; \mF_t, \mA_t)$, $\mathcal{G}_{s} =  \{\mathbf{G}_s^k, \bar{\bm{Y}}_{s}^k\}_{k=1}^K $
  \ENSURE Node classifier $h_t$ that works for $\mathbf{G}_t$
    \STATE Train node classifiers $\{ h_s^k\}_{k=1}^{K}$ for source graphs
    \STATE For each graph $\mathbf{G}_i \in \mathcal{G}_s$, train modules $\{g_T^i\}_{T\in \mathcal{T}}$ w.r.t graph modeling tasks $\mathcal{T}$ in Sec.~\ref{sec:graph_selector}
    \WHILE {Not Converged}
    \STATE Sample $\hat{\mathbf{G}}_t$, $\hat{\mathcal{G}}_s$ from $\mathcal{G}_s$
    \FOR{$\mathbf{T}$ inner update steps}
    \STATE Estimate $s_{\text{global}}$ from $\hat{\mathcal{G}}_s$ to $\hat{\mathbf{G}}_t$ with $g_{\text{sel}}^{\text{global}}$
    \STATE Estimate $s_{\text{local}}$ from $\hat{\mathcal{G}}_s$ to $\hat{\mathbf{G}}_t$ with $g_{\text{sel}}^{\text{local}}$
    \STATE Update the transport function $\gamma$ based on Eq.~\ref{eq:ot_cost} for transportation cost estimation
    \STATE Update $h_t$ following Eq.~\ref{eq:inner_update}
    \ENDFOR
    \STATE Evaluate $h_t$ for node classification on $\hat{\mathbf{G}}_t$
    \STATE Update selector parameters $\theta_{\text{sel}}$ following Eq.~\ref{eq:outer_update}
    \ENDWHILE
    \STATE Repeat line $5$ to line $10$ to update $h_t$ on the target graph $\mathbf{G}_{t}$ with source graphs $\mathcal{G}_s$
    \RETURN Trained model $h_t$ for the target graph $\mathbf{G}_{t}$
  \end{algorithmic}
\end{algorithm}

\section{Experiment}

In this section, we evaluate the proposed {\method} on $5$ real-world datasets. Specifically, we seek to answer the following questions:
\begin{itemize}[leftmargin=0.2in]
    \item \textbf{RQ1}: Can the proposed approach improve the adaptation performance on real-world graph datasets?
    \item \textbf{RQ2}: What is the importance of each component for the success of {\method}?  How would different hyperparameter configurations influence the effectiveness of {\method}?
    \item \textbf{RQ3}: Would {\method} be able to capture the importance of each source domain for this knowledge transfer? 
\end{itemize}

\subsection{Datasets}
To evaluate the effectiveness of {\method}, we conduct experiments on five publicly available datasets, including Citation~\cite{DBLP:conf/www/WuP0CZ20}, Twitch~\cite{rozemberczki2021twitch}, Yelp~\footnote{https://www.yelp.com/dataset}, Cora\_full~\cite{bojchevski2018deep} and Arxiv~\cite{hu2020open}.  We provide details of these datasets and our preprocessing in Appendix \ref{ap:dataset}.

\begin{table*}
  \caption{Multi-source graph adaptation performance
  }\label{tab:adapt_main} 
  \small
  \vskip -1em
  \begin{tabular}{c c c c c c c c c c c c }
    \toprule
    & &  & \multicolumn{5}{c}{Single-Source} &  \multicolumn{4}{c}{Multi-Source} \\
    \cmidrule(r){4-8}
    \cmidrule(r){9-12}
    Datasets & Metrics & Direct & MMD & Reverse & Adversarial & OptimalT & UDA-GCN  & DistMDA & MDAN & MLDG & {\textbf{{\method}}} \\
    \midrule
    \multirow{3}{*}{Citation} & ACC& 
     $66.5_{\scriptsize \pm 0.13}$ & 
    $62.7_{\scriptsize \pm 0.31}$ &
    $65.2_{\scriptsize \pm 1.22}$ &
    $64.8_{\scriptsize \pm 0.83}$ &
    $55.2_{\scriptsize \pm 0.69}$ &
    $65.9_{\scriptsize \pm 0.35}$ &
     $67.3_{\scriptsize \pm 0.24}$ &
    $66.5_{\scriptsize \pm 0.19}$ &
    $66.2_{\scriptsize \pm 0.12}$ &
    ${\bf{68.5}}_{\scriptsize \pm 0.22}$ 
    \\
    & AUROC & $84.1_{\scriptsize \pm 0.11}$ &
    $84.4_{\scriptsize \pm 0.29}$ &
    $85.7_{\scriptsize \pm 0.32}$ & 
    $86.1_{\scriptsize \pm 0.29}$ & 
   $79.6_{\scriptsize \pm 0.34}$ & 
   $84.7_{\scriptsize \pm 0.28}$ & 
    $84.5_{\scriptsize \pm 0.24}$ & 
   $85.5_{\scriptsize \pm 0.14}$ & 
   $85.6_{\scriptsize \pm 0.07}$ & 
    ${\bf{86.3}}_{\scriptsize \pm 0.09}$ 
    \\
    & MacroF & $64.0_{\scriptsize \pm 0.15}$ &
   $61.7_{\scriptsize \pm 0.35}$ &
   $62.2_{\scriptsize \pm 0.41}$ &
   $60.6_{\scriptsize \pm 0.33}$ &
   $48.5_{\scriptsize \pm 0.47}$ &
   $62.8_{\scriptsize \pm 0.36}$ &
   $63.2_{\scriptsize \pm 0.23}$ &
   $62.1_{\scriptsize \pm 0.12}$ &
   $62.5_{\scriptsize \pm 0.18}$ &
    ${\bf{64.8}}_{\scriptsize \pm 0.21}$ 
    
    \\
    \midrule
    \multirow{3}{*}{Twitch} & ACC& 
    $47.2_{\scriptsize \pm 0.12}$ &
    $48.3_{\scriptsize \pm 0.44}$ &
    $44.7_{\scriptsize \pm 0.0125}$ &
    $48.2_{\scriptsize \pm 0.22}$ &
    $52.2_{\scriptsize \pm 0.34}$ &
    $49.9_{\scriptsize \pm 0.47}$ &
    $51.9_{\scriptsize \pm 0.16}$ &
    $52.3_{\scriptsize \pm 0.11}$ &
    $53.6_{\scriptsize \pm 0.16}$ &
    ${\bf{57.7}}_{\scriptsize \pm 0.19}$
    \\
    & AUROC & 
   $50.9_{\scriptsize \pm 0.08}$ &
   $51.1_{\scriptsize \pm 0.19}$ & 
   $51.3_{\scriptsize \pm 0.24}$ & 
   $50.7_{\scriptsize \pm 0.16}$ & 
   $50.9_{\scriptsize \pm 0.15}$ & 
   $51.2_{\scriptsize \pm 0.16}$ & 
   $51.1_{\scriptsize \pm 0.17}$ & 
   $50.9_{\scriptsize \pm 0.16}$ & 
   $51.6_{\scriptsize \pm 0.15}$ & 
   ${\bf{51.7}}_{\scriptsize \pm 0.12}$
   \\
    & MacroF &
    $47.6_{\scriptsize \pm 0.14}$& 
   $46.7_{\scriptsize \pm 0.22}$  & 
   $44.9_{\scriptsize \pm 0.46}$ &
   $46.3_{\scriptsize \pm 0.37}$  &
   $48.3_{\scriptsize \pm 0.26}$  &
   $47.1_{\scriptsize \pm 0.31}$ & 
   $50.2_{\scriptsize \pm 0.14}$ & 
   $50.5_{\scriptsize \pm 0.08}$ &
   $50.9_{\scriptsize \pm 0.07}$ &
   ${\bf{51.5}}_{\scriptsize \pm 0.08}$
    \\
    \midrule
    \multirow{3}{*}{Yelp} & ACC& 
   ${{74.3}}_{\scriptsize \pm 0.12}$ &
    $74.6_{\scriptsize \pm 0.16}$ &
     ${{74.2}}_{\scriptsize \pm 0.09}$ &
    ${{74.1}}_{\scriptsize \pm 0.13}$ &
    ${{74.7}}_{\scriptsize \pm 0.15}$ &
    $75.9_{\scriptsize \pm 0.17}$ &
    $76.8_{\scriptsize \pm 0.18}$ &
    $78.6_{\scriptsize \pm 0.25}$ &
   ${{76.2}}_{\scriptsize \pm 0.11}$ &
   ${\bf{79.8}}_{\scriptsize \pm 0.09}$
    \\
    & AUROC & 
    ${{94.1}}_{\scriptsize \pm 0.17}$ &
   $93.5_{\scriptsize \pm 0.13}$ &
   ${{93.6}}_{\scriptsize \pm 0.11}$ &
     ${{93.0}}_{\scriptsize \pm 0.09}$ &
   ${{93.7}}_{\scriptsize \pm 0.04}$ &
   $94.1_{\scriptsize \pm 0.15}$ &
   $94.4_{\scriptsize \pm 0.12}$ & 
   $94.9_{\scriptsize \pm 0.14}$ &
   ${{94.2}}_{\scriptsize \pm 0.05}$ &
     ${\bf{95.5}}_{\scriptsize \pm 0.06}$
    \\
    & MacroF & 
     ${{74.4}}_{\scriptsize \pm 0.05}$  &
     $74.8_{\scriptsize \pm 0.12}$  &
     ${{74.0}}_{\scriptsize \pm 0.08}$ &
      ${{75.3}}_{\scriptsize \pm 0.09}$  &
      ${{74.1}}_{\scriptsize \pm 0.13}$  &
      $76.5_{\scriptsize \pm 0.16}$ &
       $77.8_{\scriptsize \pm 0.15}$ &
       $77.6_{\scriptsize \pm 0.19}$ &
        ${{77.3}}_{\scriptsize \pm 0.08}$ &
         ${\bf{79.5}}_{\scriptsize \pm 0.11}$ 
    \\
    \midrule
    \multirow{3}{*}{Cora\_full} & ACC&
     $29.4_{\scriptsize \pm 0.12}$ &
     $30.3_{\scriptsize \pm 0.12}$ &
     $31.5_{\scriptsize \pm 0.21}$ &
      $31.9_{\scriptsize \pm 0.16}$ &
    $25.4_{\scriptsize \pm 0.14}$ &
     $32.2_{\scriptsize \pm 0.17}$ &
     $31.9_{\scriptsize \pm 0.15}$ &
      $33.3_{\scriptsize \pm 0.07}$ & 
      $34.2_{\scriptsize \pm 0.15}$ &
      ${\bf{38.3}}_{\scriptsize \pm 0.19}$ 
    \\
    & AUROC & 
     $78.6_{\scriptsize \pm 0.17}$ &
    $80.1_{\scriptsize \pm 0.14}$ &
    $80.4_{\scriptsize \pm 0.19}$ &
    $80.8_{\scriptsize \pm 0.08}$ &
    $77.9_{\scriptsize \pm 0.07}$ &
     $81.0_{\scriptsize \pm 0.13}$ &
     $81.1_{\scriptsize \pm 0.11}$ &
     $81.5_{\scriptsize \pm 0.04}$ &
     $81.9_{\scriptsize \pm 0.22}$ & 
     ${\bf{83.2}}_{\scriptsize \pm 0.25}$ 
    \\
    & MacroF & 
    $19.7_{\scriptsize \pm 0.09}$  & 
     $19.9_{\scriptsize \pm 0.16}$ &  
     $20.7_{\scriptsize \pm 0.15}$&  
     $21.5_{\scriptsize \pm 0.10}$ &  
      $16.2_{\scriptsize \pm 0.13}$ &  
      $21.5_{\scriptsize \pm 0.14}$ &  
       $21.8_{\scriptsize \pm 0.16}$ &  
       $24.1_{\scriptsize \pm 0.12}$ & 
         $23.5_{\scriptsize \pm 0.13}$ &
          ${\bf{26.4}}_{\scriptsize \pm 0.08}$
    \\
    \midrule
    \multirow{3}{*}{Arxiv} & ACC& 
    $52.7_{\scriptsize \pm 0.14}$ &
    $52.6_{\scriptsize \pm 0.19}$ &
    $52.8_{\scriptsize \pm 0.14}$ &
    $53.1_{\scriptsize \pm 0.16}$ &
    $53.3_{\scriptsize \pm 0.21}$ &
    $53.4_{\scriptsize \pm 0.17}$ &
    $53.5_{\scriptsize \pm 0.21}$ &
    $53.8_{\scriptsize \pm 0.18}$ &
     $54.2_{\scriptsize \pm 0.19}$ &
      ${\bf{55.4}}_{\scriptsize \pm 0.09}$
    \\
    & AUROC &
   $91.5_{\scriptsize \pm 0.09}$ &
   $91.6_{\scriptsize \pm 0.13}$ & 
   $90.9_{\scriptsize \pm 0.13}$ & 
   $91.6_{\scriptsize \pm 0.17}$ &
   $91.9_{\scriptsize \pm 0.10}$ & 
   $91.8_{\scriptsize \pm 0.16}$ & 
   $92.0_{\scriptsize \pm 0.15}$ &
   $92.2_{\scriptsize \pm 0.14}$ &
   $91.8_{\scriptsize \pm 0.18}$ & 
    ${\bf{92.4}}_{\scriptsize \pm 0.06}$
    \\
    & MacroF &
   $27.4_{\scriptsize \pm 0.21}$&  
   $27.1_{\scriptsize \pm 0.18}$&   
   $28.9_{\scriptsize \pm 0.08}$&   
   $28.6_{\scriptsize \pm 0.14}$ &  
   $27.5_{\scriptsize \pm 0.25}$& 
   $27.9_{\scriptsize \pm 0.18}$ &   
   $29.2_{\scriptsize \pm 0.17}$ &   
   $30.1_{\scriptsize \pm 0.16}$&
   $30.4_{\scriptsize \pm 0.25}$ &
    ${\bf{31.2}}_{\scriptsize \pm 0.17}$ 
    \\
    \bottomrule
  \end{tabular}
\end{table*}

\subsection{Experiment Settings}

\subsubsection{Baselines}
The proposed method is compared with representative and state-of-the-art single-source adaptation approaches and multi-source adaptation methods. 
For empirical comparisons, we use the following baselines:
\textbf{Direct}, \textbf{MMD}~\cite{DBLP:conf/cvpr/YanDLWXZ17}, \textbf{Reverse}~\cite{DBLP:conf/icml/GaninL15}, \textbf{Adversarial}~\cite{DBLP:conf/cvpr/TzengHSD17}~\cite{DBLP:conf/www/WuP0CZ20}, \textbf{OptimalT}~\cite{DBLP:conf/nips/CourtyFHR17}, \textbf{UDA-GCN}~\cite{DBLP:conf/www/WuP0CZ20},\textbf{DistMDA}~\cite{sun2011two}, \textbf{MDAN}~\cite{zhao2018adversarial}, and \textbf{MLDG}~\cite{li2018learning}. Details are provided in Appendix \ref{ap:baseline}.

\subsubsection{Configurations}\label{sec:config}
In the experiments, for each dataset, we fix the first $5$ graphs as source domains and use the last one as the target domain, and label $10\%$ of nodes for training. 
All methods share the same backbone network structure of a two-layer GCN~\cite{kipf2016semi} and are trained until convergence with the maximum epoch number set to $2,000$. For meta-learning, the maximum epoch number of the outer update is also set to $2,000$ and the inner update step $\mathbf{T}$ is set to $5$ following~\cite{finn2017model}. The Adam optimizer is adopted for all methods, with the learning rate initialized to 0.01 and weight decay as 5e-4. For a fair comparison, for all baselines, we use grid search to set their hyperparameters. In {\method}, the hyper-parameter $\lambda$ is set to $0.3$ and $\mu$ is set to $0.001$ if not stated otherwise. 

\subsubsection{Evaluation Metrics}
We evaluate the adapted model on the target graph and adopt widely used evaluation metrics for node classification, including  accuracy (ACC), macro AUC-ROC (AUROC)~\cite{DBLP:journals/pr/Bradley97}, and macro F1 (MacroF). Both MacroAUC and MacroF are calculated and averaged per class, hence are more indicative of the existence of class imbalance.

\subsection{Graph Adaptation Performance}
To answer RQ1, we evaluate the performance of different methods after adapting to the target domain on all five datasets. Each experiment is conducted for $5$ times with random parameter initialization and train/test splits of source domains, and we report the average results in Table~\ref{tab:adapt_main}. Both the mean and standard deviation w.r.t each metric are presented, with the best performance emboldened. From the results, we can make the following observations:
\begin{itemize}[leftmargin=0.1in]
    \item Our proposed method, {\method}, achieves the best performance across all these datasets and outperforms baselines with a clear margin. For example, on dataset  Citation and Twitch, it shows an improvement of $1.2\%$ and $4.1\%$ in terms of accuracy compared to the best baseline. 
    \item Adaptation by taking all source domains as equal and neglecting their varying transferability to the target domain often shows weak performances, and may even result in a performance drop compared to no adaptation at all. For example, on dataset Citation, all single-source domain alignment methods show a performance drop in both accuracy and Macro F score compared to the baseline Direct, which trains a global source model and directly applies it to the target domain.
    \item Compared to other multi-source adaptation methods like DistMDA, \method achieves further improvement, which supports our motivation to capture transferability between graph domains to cope with the highly diverse topology structure.
\end{itemize}

These results validate the effectiveness of {\method}, showing its strong performance in the multi-source graph adaptation task.

\subsection{Ablation Study}
In order to examine the importance of each component and answer RQ2, in this subsection, we conduct a set of ablation studies by removing different parts of {\method} and test the performance. All hyper-parameters are left unchanged, and each experiment is conducted three times on datasets Twitch, Yelp, and Cora\_full with random initialization and train/test splits of source domains. Mean performance and standard deviations are summarized in Table~\ref{tab:ablation}.

\begin{table*}
  \caption{Ablation Study on each framework component.  }\label{tab:ablation}
  \small
  \vskip -1em
  \begin{tabular}{c c c c c c c c c c }
    \toprule
    &    \multicolumn{3}{c}{Twitch} &  \multicolumn{3}{c}{Yelp}   & \multicolumn{3}{c}{Cora\_full} \\
    \cmidrule(r){2-4}
    \cmidrule(r){5-7}
    \cmidrule(r){8-10}
    Methods & ACC & AUROC & MacroF & ACC & AUROC & MacroF & ACC & AUROC & MacroF \\
    \midrule
    {\method} &${\bf{57.7}}_{\scriptsize \pm 0.19}$ & ${\bf{51.7}}_{\scriptsize \pm 0.12}$ & ${\bf{51.5}}_{\scriptsize \pm 0.08}$ & ${\bf{79.8}}_{\scriptsize \pm 0.09}$ & ${\bf{95.5}}_{\scriptsize \pm 0.06}$ & ${\bf{79.5}}_{\scriptsize \pm 0.11}$ &${\bf{38.3}}_{\scriptsize \pm 0.19}$ & ${\bf{83.2}}_{\scriptsize \pm 0.25}$ & ${\bf{26.4}}_{\scriptsize \pm 0.08}$\\
    w/o $g_{\text{sel}}^{\text{global}}$ & 
     $54.3_{\scriptsize \pm 0.13}$ &
   $51.3_{\scriptsize \pm 0.15}$ & 
   $50.9_{\scriptsize \pm 0.09}$
    & $77.0_{\scriptsize \pm 0.11}$ & $93.6_{\scriptsize \pm 0.08}$ & $77.5_{\scriptsize \pm 0.14}$ & $35.5_{\scriptsize \pm 0.25}$ & $82.8_{\scriptsize \pm 0.22}$ & $25.0_{\scriptsize \pm 0.16}$ \\
    w/o $g_{\text{sel}}^{\text{local}}$&
    $55.2_{\scriptsize \pm 0.16}$ & $51.5_{\scriptsize \pm 0.14}$  & $51.0_{\scriptsize \pm 0.12}$
    & $77.7_{\scriptsize \pm 0.12}$ & $94.7_{\scriptsize \pm 0.13}$  & $77.8_{\scriptsize \pm 0.12}$   & $35.7_{\scriptsize \pm 0.31}$ & $82.6_{\scriptsize \pm 0.13}$ & $25.4_{\scriptsize \pm 0.28}$ \\
    w/o KD & 
    $53.7_{\scriptsize \pm 0.16}$ & $51.4_{\scriptsize \pm 0.15}$  & $51.1_{\scriptsize \pm 0.10}$
    & $78.9_{\scriptsize \pm 0.10}$ & $95.1_{\scriptsize \pm 0.11}$ & $78.1_{\scriptsize \pm 0.16}$ & $36.5_{\scriptsize \pm 0.23}$ & $83.0_{\scriptsize \pm 0.19}$ & $25.7_{\scriptsize \pm 0.13}$ \\
    SelAdv &
    $55.3_{\scriptsize \pm 0.13}$ &
   $51.6_{\scriptsize \pm 0.25}$ & 
   $51.4_{\scriptsize \pm 0.19}$
    & $79.5_{\scriptsize \pm 0.13}$ & $95.2_{\scriptsize \pm 0.12}$ & $79.2_{\scriptsize \pm 0.09}$ & $36.8_{\scriptsize \pm 0.22}$ & $83.1_{\scriptsize \pm 0.09}$ & $25.9_{\scriptsize \pm 0.29}$ \\
    \bottomrule
\end{tabular}
\vskip -1em
\end{table*}

\subsubsection{Source-Graph Selectors}

To evaluate the importance of different weights assigned to source domains, we test the performance after removing model-based graph selector (w/o $g_{\text{sel}}^{\text{global}}$ ) and sub-graph node selector (w/o $g_{\text{sel}}^{\text{local}}$). These two variants correspond to removing $s_{\text{global}}$ and $s_{\text{local}}$ in Eq.~\ref{eq:weight}, respectively. From Table~\ref{tab:ablation}, it is shown that both selectors play a positive influence in this multi-source adaptation. Particularly, the model-based graph selector is more important than the sub-graph node selector in most cases, with the variant w/o $g_{\text{sel}}^{\text{global}}$ showing the worst accuracy in all three datasets. The reason could lie in the utilization of graph modeling tasks which provides clearer evidence for estimating transferability.

\subsubsection{Domain Alignment Objectives}\label{sec:ab_alignment}
To evaluate the importance of aligning the classification function, we implement a variant by removing the weighted knowledge distillation loss $\mathcal{L}_{\text{KD}}$ in Eq.~\ref{eq:full_obj} and annotate it as ``w/o KD''. To test the generalizability of {\method} with different domain alignment objectives, we further implement a variant SelAdv by replacing OT-based alignment in $\mathcal{L}_{\text{SelOT}}$ with adversary-based alignment~\cite{zhao2018adversarial}. Learned weights of source domains can be incorporated into the adversarial aligning process via re-weighting. From Table~\ref{tab:ablation}, it can be observed that the weighted knowledge distillation loss is also helpful for the adaptation process, with ``w/o KD'' showing a drop in adaptation performance. Besides, it is shown that our method can also work for the adversarial alignment with a moderate performance drop, validating its robustness across alignment strategies.

\subsection{Sensitivity Analysis}
\subsubsection{Alignment Weight}
\begin{figure}[t!]
  \centering
  \begin{subfigure}{0.235\textwidth}{
		\includegraphics[width=\textwidth]{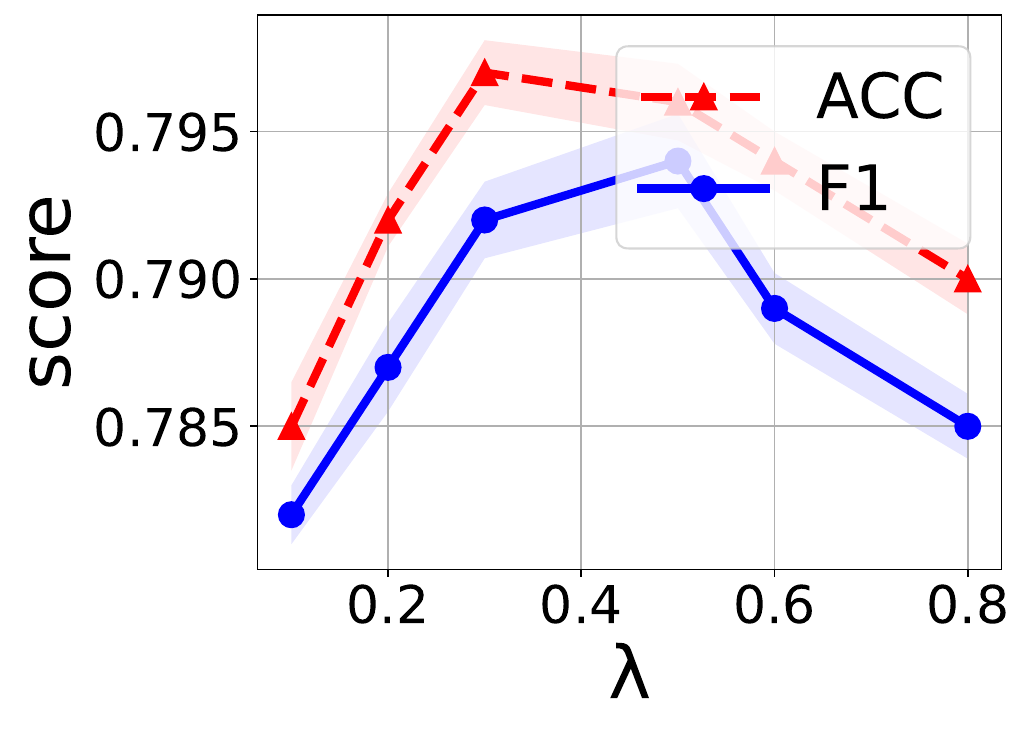}}
  \vskip -1em
  \caption{Yelp}
  \end{subfigure}
  \begin{subfigure}{0.23\textwidth}{
		\includegraphics[width=\textwidth]{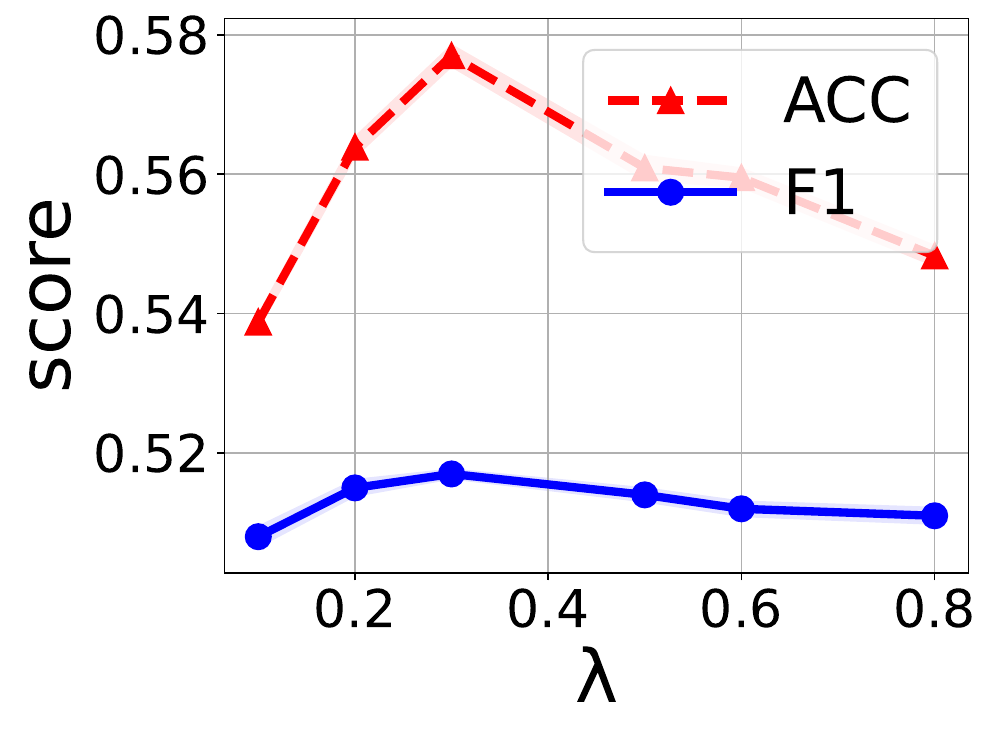}}
  \vskip -1em
  \caption{Twitch}
  \end{subfigure}
    \vskip -1em
    \caption{Influence of the weight of domain alignment loss. } \label{fig:lam_1}
    \vskip -1.5em
\end{figure}

To evaluate the domain adaptation objective and partially answer RQ2, we analyze the sensitivity of the {\method} on hyper-parameters $\lambda$, which controls the balance between the embedding alignment loss $\mathcal{L}_{\text{SelOT}}$ and the knowledge distillation loss $\mathcal{L}_{\text{KD}}$ in Eq.~\ref{eq:full_obj}. We vary it as $\{0.1, 0.2, 0.3, 0.5, 0.6, 0.8\}$, and experiments are conducted on Twitch and Yelp, with other configurations remaining the same as the main experiment. Each experiment is conducted $3$ times, and we report the average accuracy and Macro F score in Fig.~\ref{fig:lam_1}. It is shown that $\lambda$ is better set to the range $[0.3, 0.6]$ for both datasets. Setting it too small would weaken the embedding alignment signal while setting it too high would fail to explicitly align the classifier.

\subsubsection{Optimal Transport Configuration}
\begin{figure}[t!]
  \centering
  \begin{subfigure}{0.235\textwidth}{
		\includegraphics[width=\textwidth]{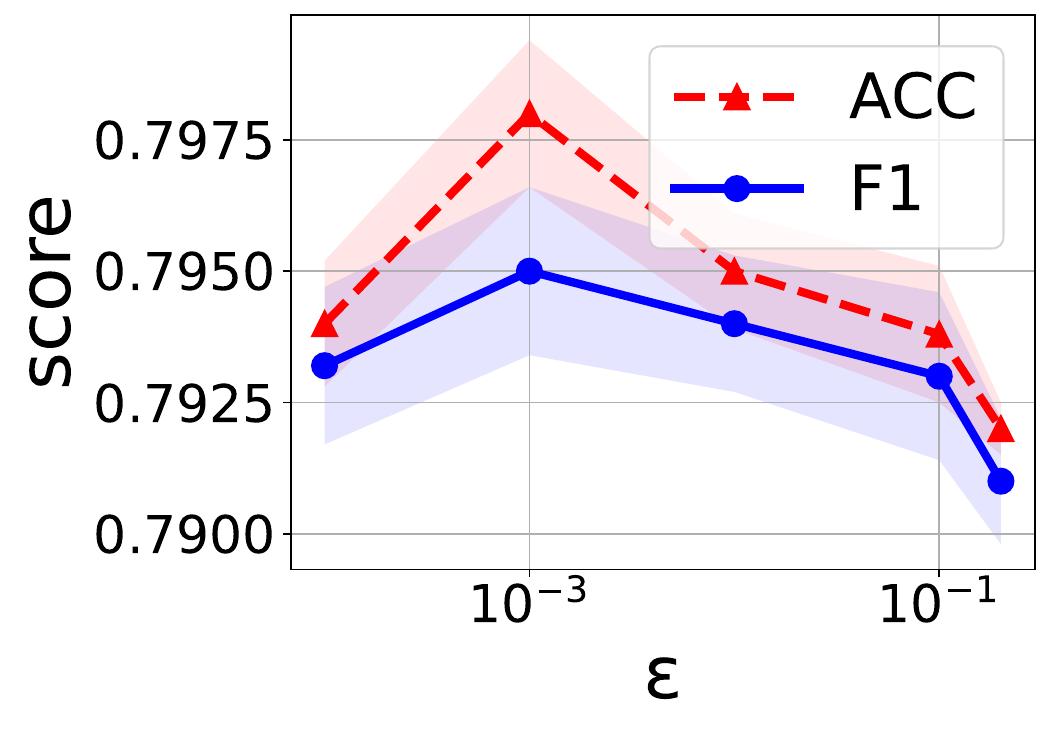}}
  \vskip -1em
  \caption{Yelp}
  \end{subfigure}
  \begin{subfigure}{0.22\textwidth}{
		\includegraphics[width=\textwidth]{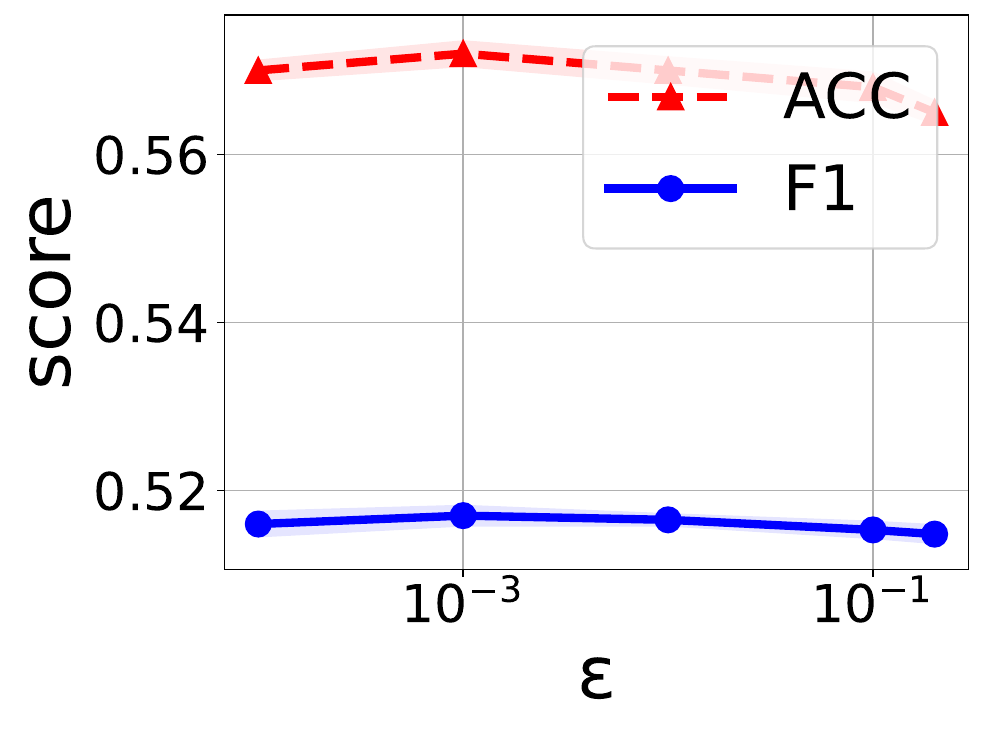}}
  \vskip -1em
  \caption{Twitch}
  \end{subfigure}
    \vskip -1em
    \caption{Influence of the hyper-parameter $\epsilon$, which controls optimal transport in Eq.~\ref{eq:ot_cost}. } \label{fig:lam_2}
    \vskip -1.5em
\end{figure}

To analyze the sensitivity of {\method} on the optimal transport configuration, in this part we vary $\epsilon$ as $\{0.0001, 0.001, 0.01, 0.1, 0.2\}$ which regularizes the transport in Eq.~\ref{eq:ot_cost}. A larger $\epsilon$ encourages the transport plan to be smoother~\cite{nguyen2021most}. Again, experiments are conducted on Twitch and Yelp for $3$ times with all other configurations unchanged. The average results are presented in Fig~\ref{fig:lam_2}. It is shown that {\method} achieves relatively stable performance with a small $\epsilon$ within $[0, 0.1]$, in accordance with previous observations~\cite{genevay2016stochastic}.

\subsection{Analysis on Global and Local Weights}\label{sec:case}
\begin{figure}[t]
  \centering
  \begin{subfigure}{0.235\textwidth}{
		\includegraphics[width=\textwidth]{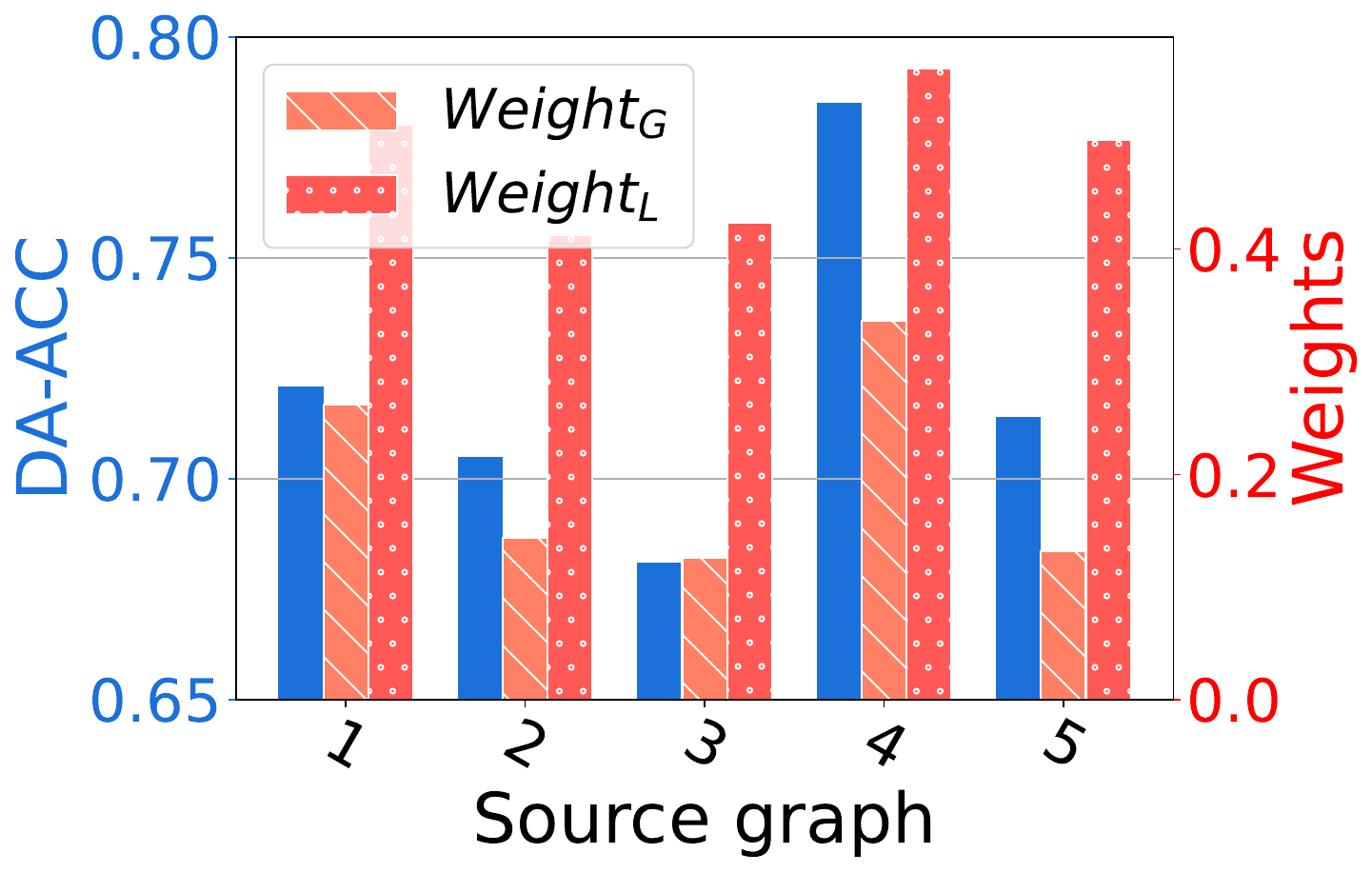}}
  \vskip -0.5em
  \caption{Yelp}
  \end{subfigure}
  \begin{subfigure}{0.235\textwidth}{
		\includegraphics[width=\textwidth]{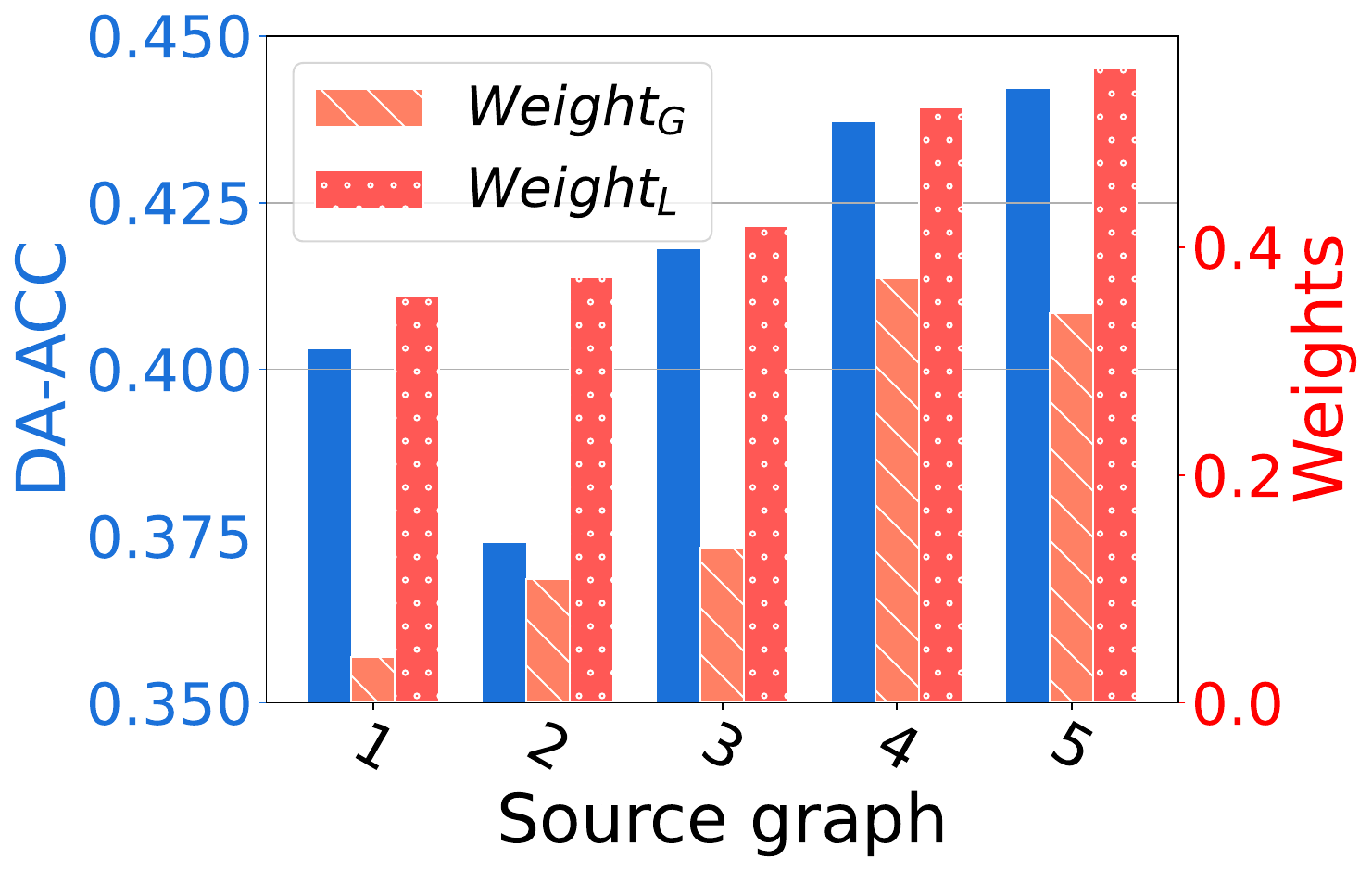}}
  \vskip -0.5em
  \caption{Citation}
  \end{subfigure}
    \vskip -1em
    \caption{Analysis on weights assigned to source domains by $g_{\text{sel}}^{\text{global}}$ and $g_{\text{sel}}^{\text{local}}$, by comparing them to the accuracy obtained with single-source domain adaptation.} \label{fig:case}
    \vskip -1.5em
\end{figure}

In this subsection, we provide some analysis over learned weights $s_{\text{global}}$ and $s_{\text{local}}$ of source graph domains towards the target graph in order to answer RQ3. However, there is no ground-truth transferability between graph pairs w.r.t the downstream node classification task. Therefore, we take a surrogate strategy by evaluating the single-source graph adaptation performance. Single-source graph adaptation is conducted by taking only one source graph as available and using optimal transport for embedding alignment. The accuracy obtained on the target domain after adaptation is reported as DA-ACC. Experiments are conducted on Yelp and Citation, with optimization hyper-parameters set to be the same as introduced in Sec.~\ref{sec:config}. For ease of analysis, we apply only the graph selector or the sub-graph node selector respectively to exclude the influence of the other one. Results are summarized in Fig.~\ref{fig:case}. For $s_{\text{local}}$, we report its mean across nodes of the corresponding domain.


From Fig.~\ref{fig:case}, we can observe that {\method} tends to assign a higher weight to source domains that show a strong performance in single-source UDA. For example on Yelp, the largest global and local weights are both assigned to $4$-th graph, which also shows the strongest UDA performance. On Citation, the $5$-th source graph is generated from the same dataset as $\mathbf{G}_t$, and its weights are also high. For important source domains, their average local node weights are also high. These results validate the ability of {\method} in estimating transferability between graph domains.

\section{Conclusion}

{In this work, we propose a novel framework to identify informative sub-graphs for knowledge transfer in the multi-source graph adaptation task. We depict graph similarities from three perspectives, each captured by a self-supervised graph modeling task, and estimate task-specific cross-domain transferability with a meta-learned selector. An optimal-transport objective and a weighted knowledge distillation objective are designed to incorporate obtained selection scores into the domain alignment process. } Experiments on five datasets show that 
{\method} outperforms existing methods for MSDA on graphs. In the future, we plan to extend {\method} to work with emerging new classes. In adapting source models to the target domain, there could be shifts in the label space, like novel classes unseen during training. This scenario calls for the development of a new algorithm.    


\section{Acknowledgement}
This material is based upon work supported by, or in part by the National Science Foundation (NSF) under grant number IIS-1909702, Army Research Office (ARO) under grant number W911NF-21-1-0198, Department of Homeland Security (DHS) CINA under grant number E205949D, and Cisco Faculty Research Award.

\bibliographystyle{ACM-Reference-Format}
\balance
\bibliography{file1}

\clearpage

\appendix


\section{Dataset Description}\label{ap:dataset}
\begin{itemize}[leftmargin=0.1in]
\item \textbf{Citation}. In this dataset, we use three citation networks collected from ACM (ACM-V9), DBLP (DBLP-V7), and Microsoft Academic Graph (Citation-V1) respectively. Each node represents a paper and its descriptions are extracted as attributes using Bog-of-words. Edges denote citations, and nodes are labeled on the paper domain. We randomly split each network into two graphs and create a dataset of $6$ graphs, which can increase the graph number and provide us with some prior knowledge on the similarity between graph pairs at the same time. This design can also help evaluate the effectiveness of our framework in selecting source domains. A sub-graph of Citation-V1 is used as the target.
\item \textbf{Twitch}. This dataset is collected from the Twitch gamer platform, with nodes as Twitch users and edges as mutual follower relationships between them. This binary node classification task predicts whether a user streams explicit content. Six graphs are obtained based on the language used by the user: French, Spanish, Portuguese, German, English or Russian. The graph with language Russian is used as the target graph.
\item \textbf{Yelp}. This dataset contains user reviews on Yelp to various point-of-interests (POIs) in different cities. We transform the reviews in each city into a graph, with each node representing a POI and each edge representing a co-review relationship. POI features are obtained by averaging the word embedding of its reviews, which is taken from the pre-trained language model GLOVE~\cite{Pennington2014GloveGV}. We perform classification on five classes, \{\textit{Food, Shop, Home Service, Health Service, Finance}\}, and select six cities of different scales: \{\textit{Madison, Glendale, Gilbert, Las Vegas, Toronto, Phoenix}\}. City Phoenix is used as the target graph.
\item \textbf{Cora\_full}. It is a citation network with nodes for papers and edges for citation relations. We cluster papers into $6$ groups based on different frequencies in selected-words usage following GOOD~\cite{guigood} to generate $6$ graphs and select one as the target.
\item \textbf{Arxiv}. It is a large citation network among the computer science (CS) arXiv papers. The task is to predict the subject area of each paper. We split it into $6$ disjoint graphs based on the published year of each paper~\cite{guigood}, and use the most recent one as $\mathbf{G}_t$.
\end{itemize}
The main statistics of these datasets are summarized in Table~\ref{tab:dataset}, including average node numbers and average edge numbers among graphs of each dataset.

\begin{table}[h]
  \caption{Statistics of processed graph datasets. }\label{tab:dataset}
  \vskip -1em
  \small
  \centering
  \begin{tabular}{cccccc}
    \toprule
    Name     & \#Graphs & \#Nodes & \#Edges & \#Attributes & \#Classes \\
    \midrule
    Citation & $6$  & $3,963$ & $6,428$ & $6,775$ & $5$     \\
    Twitch   & $6$  & $5,687$ & $148,724$ & $128$ & $2$    \\
    Yelp     & $6$  & $13,903$ & $232,525$ & $100$ & $6$ \\
    \midrule
    Cora\_full & $6$ & $3,298$ & $3,697$  & $8,710$ & $70$  \\
    Arxiv &  $6$ & $28,223$ & $66,166$ & $128$ & $40$ \\
    \bottomrule
  \end{tabular}
  \vskip -1em
\end{table}

\section{Baseline Introduction}\label{ap:baseline}
\begin{itemize}[leftmargin=0.1in]
    \item \textbf{Direct}. Instead of conducting explicit alignment in this baseline, we train a global model by using all the source domains and directly apply it to the target graph. 
    \item \textbf{MMD}~\cite{DBLP:conf/cvpr/YanDLWXZ17}. This baseline conducts cross-domain alignment by minimizing the maximum mean discrepancy (MMD) of embedding distributions between the source and target graphs.
    \item \textbf{Reverse}~\cite{DBLP:conf/icml/GaninL15}. By augmenting feed-forward models with a simple new gradient reversal layer, this baseline can encourage discovering features that are not predictive towards domains.
    \item \textbf{Adversarial}~\cite{DBLP:conf/cvpr/TzengHSD17}~\cite{DBLP:conf/www/WuP0CZ20}. To capture generalizable features that are invariant across domains, this baseline adversarially trains a domain discriminator to distinguish the source and target domains. Model is learned to extract features that are both discriminative towards node labels and can fool the domain discriminator.
    \item \textbf{OptimalT}~\cite{DBLP:conf/nips/CourtyFHR17}. This baseline assumes a non-linear transformation between the joint feature/label space distributions across domains, and the model is adapted by minimizing this total transformation cost.
    \item \textbf{UDA-GCN}~\cite{DBLP:conf/www/WuP0CZ20}: This method aligns source and target graphs with a domain classifier and includes classification entropy to promote classification boundaries of the target domain.
    \item \textbf{DistMDA}~\cite{sun2011two}. This method is designed for multi-source unsupervised domain adaptation, which makes a smoothness assumption on data distribution and estimates the weight of each source domain by minimizing the marginal probability difference. After obtaining source weights, Optimal transport is used for alignment. We extend it to graph-structured data by computing probability differences in the embedding space.
    \item \textbf{MDAN}~\cite{zhao2018adversarial}. A worst-case selection strategy is used in this work for multi-source unsupervised domain adaptation. We use its Soft-Max version to assign weights and then conduct OT-based adaptation.
    \item \textbf{MLDG}~\cite{li2018learning}. Meta-learning with synthesized virtual testing domains is utilized in this method in order to explicitly learn to generalize to the unseen domain with multiple source domains available. We implement it in the same setting as our {\method}, with MAML~\cite{finn2017model} for the conduction of meta-learning part.
\end{itemize}
Note that MMD, Reverse, Adversarial, OptimalT, and UDA-GCN are originally designed for single-source domain adaptation. We extend them to MSUDA by taking all source domains as one, equivalent to giving the same weights to all source graphs.

\section{Time Complexity}
{In this part, we provide some analysis over the additional computation cost of our method. Note that during model adaptation process, at each batch, both baselines and our method have the same backbone GNN architectures and the domain distance estimator for alignment. In addition to these backbone modules, our algorithm needs an additional source domain selector. Denoting number of GNN layers as $L$, number of edges in the dataset as $E$, number of nodes as $n$ and embedding dimension as $d$, for the backbone model, time complexity of GNN is $O(LEd+Lnd^2)$ and time complexity of distance estimator is $O(nd^2)$. For our model, the additional domain selector has the time complexity of $O(nd^2)$. Therefore, the proposed $\method$ will not result in a significant computation increase in adaptation.
During learning, the time-consuming part of our model is to pretrain the graph modeling tasks.  For a concrete example, on dataset Arxiv, which has 169,338 nodes, the pretraining step takes around one hour. Note that this pretraining step is only needed once across different running, hence its cost would not pose a severe problem. }

\end{document}